\documentclass{ecai} 
\usepackage{latexsym}
\usepackage{amssymb}
\usepackage{amsmath}
\usepackage{amsthm}
\usepackage{enumitem}
\usepackage{graphicx}
\usepackage{color}
\usepackage{algorithm}
\usepackage{algpseudocode}
\usepackage{xspace}
\usepackage{amsmath} 
\usepackage{amsfonts}
\usepackage{lscape}

\usepackage{booktabs}
\usepackage{multirow}
\usepackage{multicol}
\usepackage{array}
\usepackage{makecell}

\usepackage{enumitem}
\usepackage{svg} % Include the svg package
\usepackage{color}
\usepackage{xcolor}
\usepackage{tcolorbox}
\newcolumntype{C}[1]{>{\centering\arraybackslash}p{#1}}
\usepackage{algorithm}
\usepackage{algpseudocode}
\usepackage{enumitem}
\usepackage{subcaption}
\usepackage{xspace}
\usepackage{xr}
\usepackage{hyperref}

\newcommand{\BibTeX}{B\kern-.05em{\sc i\kern-.025em b}\kern-.08em\TeX}

\begin{document}

%%%%%%%%%%%%%%%%%%%%%%%%%%%%%%%%%%%%%%%%%%%%%%%%%%%%%%%%%%%%%%%%%%%%%%%%

\begin{frontmatter}

%%% Use this command to specify your submission number.
%%% In doubleblind mode, it will be printed on the first page.

\paperid{2301} 

%%% Use this command to specify the title of your paper.

\title{Navigating Trade-offs: Policy Summarization for Multi-Objective Reinforcement Learning}

%%% Use this combinations of commands to specify all authors of your 
%%% paper. Use \fnms{} and \snm{} to indicate everyone's first names 
%%% and surname. This will help the publisher with indexing the 
%%% proceedings. Please use a reasonable approximation in case your 
%%% name does not neatly split into "first names" and "surname".
%%% Specifying your ORCID digital identifier is optional. 
%%% Use the \thanks{} command to indicate one or more corresponding 
%%% authors and their email address(es). If so desired, you can specify
%%% author contributions using the \footnote{} command.

\author[A]{\fnms{Zuzanna}~\snm{Osika}\orcid{0000-0002-0602-2812}\thanks{Corresponding Author. Email: z.osika@tudelft.nl}}
\author[A]{\fnms{Jazmin}~\snm{Zatarain-Salazar}\orcid{0000-0003-2248-2242}}
\author[A]{\fnms{Frans A.}~\snm{Oliehoek}\orcid{0000-0003-4372-5055}} 
\author[A]{\fnms{Pradeep K.}~\snm{Murukannaiah}\orcid{0000-0002-1261-6908}}

\address[A]{Delft University of Technology}

%%% Use this environment to include an abstract of your paper.
\begin{abstract}
Multi-objective reinforcement learning (MORL) is used to solve problems involving multiple objectives. An MORL agent must make decisions based on the diverse signals provided by distinct reward functions. Training an MORL agent yields a set of solutions (policies), each presenting distinct trade-offs among the objectives (expected returns). MORL enhances explainability by enabling fine-grained comparisons of policies in the solution set based on their trade-offs as opposed to having a single policy. However, the solution set is typically large and multi-dimensional, where each policy (e.g., a neural network) is represented by its objective values.

We propose an approach for clustering the solution set generated by MORL. By considering both policy behavior and objective values, our clustering method can reveal the relationship between policy behaviors and regions in the objective space. This approach can enable decision makers (DMs) to identify overarching trends and insights in the solution set rather than examining each policy individually. We tested our method in four multi-objective environments and found it outperformed traditional k-medoids clustering. Additionally, we include a case study that demonstrates its real-world application.

\end{abstract}
\end{frontmatter}

\section{Introduction}

% Multi-Objective Reinforcement Learning - what is it
Multi-objective reinforcement learning (MORL) is a branch of reinforcement learning (RL) that aims to optimize multiple, often conflicting, objectives simultaneously. MORL has been applied in various domains, including water management (\cite{casteletti-momd-2012,giuliani2016curses}), autonomous driving (\cite{li-urban-2019}), power allocation (\cite{oh2023multi,xiong2023multi}), drone navigation (\cite{wu2024multi}), and medical treatment (\cite{jalalimanesh2017multi,lizotte-2010-efficient}). 
Unlike single objective RL that focuses on maximizing a single reward function, MORL requires balancing trade-offs among multiple goals. As a result, the output of an MORL algorithm is not a single policy, but a set of policies. Depending on the assumptions and the information available, the solution set can be a Pareto set, a coverage set, or a convex hull \cite{hayes2022practical}.
%

% Explainability in MORL
A significant advantage of MORL is that it is inherently more explainable than single-objective RL. MORL outputs a rich set of policies (solutions), each offering distinct trade-offs with regard to the objectives. This multiplicity of solutions not only provides a comprehensive overview of the possible actions and their consequences but also facilitates a deeper understanding of how objectives interact and influence decision-making.  Further, MORL avoids the need to collapse multiple goals into a single metric. Thus, decision makers (DMs) could be equipped with a nuanced and detailed understanding of the outcomes, thereby enabling them to make informed choices. This transparency in policy generation enhances the trustworthiness of decisions made using MORL in real-world problems.

% Bottlenecks of MORL - hard to understand the trade-offs
Although MORL makes trade-offs explicit, the DMs must still exert considerable cognitive effort to navigate and evaluate the options in the solution set. This process becomes more challenging if the DMs do not have clear preferences on the objectives. Further, as the number of objectives increases, the size of the solution set expands drastically, which makes comprehending the entire set of solutions impractical. Thus, although MORL holds considerable promise in enhancing the explainability of decisions, its practical application is hindered by the associated complexities. To fully leverage MORL's potential, it is important to develop methodologies and tools to support the DM in exploring the policies in the solution set. %Such advancements are crucial to transform the theoretical benefits of MORL into tangible decision-making support.

% Current state of post-analysis MORL (none)
The main focus of current MORL research is on developing algorithms and benchmarking their performance (e.g., \cite{alegre2023gpi,felten_toolkit_2023,xu2020prediction}). However, there is a dearth of tools that allow users (DMs) to comprehend the high-dimensional solution sets the MORL algorithms produce. Such tools facilitate the practical deployment of MORL systems by enabling DMs to analyze policy behavior, understand the trade-offs, and differentiate between policies. Without these decision-support tools, there is a gap in the practical applicability of MORL, hindering the translation of algorithmic advancements into real-world applications. To address this gap, we propose approaches for the post-analysis of trade-offs in the MORL solution sets.

% Clustering as one of the popular methods
In one-shot multi-objective optimization (MOO), e.g., engine design, a common method to convey a high-dimensional solution set involves clustering the solutions based on their trade-offs in the objective space (see \cite{Osika-2023-IJCAISurveys-MODMDecisionSupport} for a review on decision support approaches for traditional one-shot MOO). Clustering of solutions \cite{BANDARU2017139} in the objective space can streamline the decision-making process by reducing the dimensionality of the solution set, enabling DMs to focus on clusters of solutions rather than evaluating all solutions. Ulrich \cite{ulrich-biobjective-2013} extends the standard clustering approach by proposing Pareto-Set Analysis, an approach that identifies compact and well-separated clusters in both decision and objective spaces. This approach aids in identifying trends and key turning points in the data and it reveals the relationships between decision variables and objectives.

% Extending the clustering approaches to morl setting
We extend the idea of Pareto-Set Analysis to MORL solution sets. Whereas clustering in the objective space is similar to one-shot MOO, finding clusters in the decision variable space is nontrivial for MORL. In MORL, a solution, policy, is described as a mapping from states to a probability distribution over actions. For problems with large (sometimes infinite) state or action spaces, the solution can be a neural network, making post-analysis in both spaces challenging.

Clustering in the objective space alone is inadequate, as different policies with similar trade-offs can behave differently. For example, two water management policies may yield similar expected returns but differ drastically in terms of disaster risk. Therefore, it's crucial to summarize and compare policy behaviors relative to their expected returns to understand how they align with or diverge from objectives.

\paragraph{Contribution.}
We address the challenge of decision support in MORL by proposing a method that explains the MORL solution set through clustering, considering both the objective and the behavior spaces. To do so, we show ways to effectively represent the behavior space of the policies by employing techniques from explainable reinforcement learning (XRL) field. Given that representation, we apply Pareto-Set Analysis (PAN) clustering algorithm \cite{ulrich-biobjective-2013} to find well-defined clusters in both spaces. We apply this method to four diverse MORL environments and show that it improves performance, compared to the traditional k-medoid clustering. We also use one of the environments, the multi-objective highway environment, to demonstrate the importance of considering not just the trade-offs (expected returns) but also policies' behavior when presenting information to the DMs. To the best of our knowledge, we are the first to tackle the challenge of explaining the solution set of MORL. \footnote{The code is available at \href{https://github.com/osikazuzanna/Bi-Objective-Clustering}{https://github.com/osikazuzanna/Bi-Objective-Clustering}}

\section{Background and Related Work}

We review three lines of work we build on: 1) MORL, 2) Pareto Set Analysis, and
3) policy summarization.

\subsection{Multi-Objective Reinforcement Learning (MORL)}

MORL is used to solve problems where an agent should optimize multiple, potentially conflicting, objectives simultaneously over time. Each objective in MORL is represented by a separate reward function. These problems can be modeled as Multi-objective Markov Decision Processes (MOMDPs) with multiple reward functions. 

In MORL, a MOMDP is defined by \( (S, A, p, r, \mu, \gamma) \), where:

\begin{itemize}
    \item \(S \) represents the state space.
    \item \(A \) is the action space.
    \item \( p(\cdot \mid s, a) \) is the probability distribution over next states given the current state s and action a.
    \item \( r: S \times A \times S \rightarrow \mathbb{R}^m \) is the multi-objective reward function with $m$ objectives.
    \item \( \mu \) is the initial state distribution.
    \item \(\gamma \in [0, 1) \) is a discount factor.
\end{itemize}

An MORL policy, \( \pi: S \rightarrow A \), maps states to actions to optimize outcomes across objectives. The action-value function for a policy \(\pi\) for a given state-action pair \( (s, a) \) is an m-dimensional vector representing the expected return for each objective \cite{hayes2022practical}.

A key concept in MORL is the Pareto frontier---a set of policies whose multi-objective value functions are not dominated by any other policy. A policy's value profile is said to be Pareto dominant if it is better or equal in all objectives and strictly better in at least one objective compared to another policy's value function. The Pareto frontier thus consists of all policies that offer the best possible trade-offs among the objectives, where improving one objective cannot be achieved without worsening at least one other objective \cite{moffaert2013morl}.

Another key concept is the Convex Hull, specifically in the context of linear utility functions. It defines the region where the weighted sum of objectives (as determined by the utility function) can be maximized. For a set of weights, which represent the relative importance of each objective, the optimal policy is one that maximizes this weighted sum. The Convex Hull, therefore, contains policies that are optimal for different combinations of weights, covering all linear preferences in the multi-objective framework \cite{roijers2013survey}.

Hypervolume is a popular metric for measuring the performance of MORL algorithms. It measures the volume (in the objective space) enclosed between the solution set generated by an algorithm and a reference point, which typically represents the worst possible values of objectives \cite{Wang2013HypervolumeIA}. The metric provides a measure of how well a set of solutions covers the objective space, indicating the diversity and quality of the solutions with respect to the multiple objectives being optimized. A larger hypervolume generally indicates a diverse set which is also close to the ideal solution.

\textbf{GPI-PD.} In our experiments, we employ the Generalized Policy Improvement-Prioritized Dyna (GPI-PD) algorithm \cite{gpipdalegre23} to train our agents. 
%General overview
The algorithm leverages Generalized Policy Improvement (GPI) to establish prioritization schemes that enhance the efficiency of learning from samples. Through the utilization of active learning strategies, the algorithm enables the agent to (i) pinpoint the most promising objectives or preferences to focus on at any given time, accelerating the resolution of MORL challenges; and (ii) determine the most pertinent previous experiences to inform the learning of a policy tailored to a specific agent preference, utilizing an innovative approach inspired by Dyna-style MORL methods. 

%GPI
Generalized Policy Improvement (GPI) extends policy improvement by defining a new policy \(\pi\) that improves over a set of policies \(\Pi\), rather than a single one. For a given weight vector \(\mathbf{w} \in \mathcal{W}\), the GPI policy is \(\pi^{GPI}(s; \mathbf{w}) \in \arg\max_{a \in \mathcal{A}} \max_{\pi \in \Pi} q^{\pi}_{\mathbf{w}}(s, a)\). GPI ensures that the selected policy matches or surpasses the performance of any policy \(\pi_i \in \Pi\) across all weight vectors \(\mathbf{w} \in \mathcal{W}\).

%Dyna
Dyna-style algorithms use a search control mechanism to sample experiences from the learned model for planning. They construct a set of policies \(\Pi\) with value vectors \(V\) approximating the CCS. Each iteration involves selecting a weight vector \(\mathbf{w} \in \mathcal{W}\), guided by GPI improvements, and learning a new policy \(\pi_w\) optimized for that weight vector. The algorithm employs GPI-based prioritization to efficiently determine relevant experiences for learning the optimal policy \(\pi_w\). GPI-PD learns a multi-objective dynamics model \(p\) to predict subsequent states and rewards from state-action pairs, facilitating Dyna updates to the action-value functions using simulated experiences.

\subsection{Pareto-Set Analysis}

Pareto-Set Analysis (PAN) is a clustering technique introduced in the field of MOO \cite{ulrich-biobjective-2013}. Unlike the traditional clustering approaches that focus solely on the objective space (e.g. k-means clustering), PAN identifies clusters that are compact and well-separated in both decision and objective spaces. Also, PAN does not focus on finding just one best partitioning of the solutions. Instead, it helps the DM to better understand the problem by revealing the relationship between different design choices and their outcomes in the objective space.

PAN is flexible as it does not depend on the identification of specific design variables or feature vectors. The main prerequisite is a distance measure in each space. Since the clusters formed in the decision space may not align with those in the objective space, PAN formulates the clustering as a biobjective optimization challenge (partitioning to perform well in both objective and decision spaces). Thus, the PAN algorithm is a bi-objective evolutionary algorithm tailored for generating a set of effective partitionings.

The essence of PAN is its ability to optimize partitioning in both decision and objective spaces, utilizing two validity indices of the clustering: one for the objective space, \(V(f(C^*), d_O)\), and another for the decision space, \(V(f(C^*), d_D)\). These indices evaluate the coherence of solutions within each cluster, guiding the algorithm toward partitionings that optimize the trade-offs between being compact in one space and well-separated in the other. 

PAN applies variation and selection processes, iteratively, to evolve a population of partitionings. Each partitioning is assessed based on its clustering quality in both spaces, marking a significant departure from conventional methods that prioritize objectives without considering the structural implications of solutions.

\subsection{Policy Summarization Techniques in RL}

% Importance of explainability

We aim to adapt PAN clustering to MORL policy set. Unlike traditional one-shot MOO tasks that focus on one-time decisions, MORL involves sequential decision-making processes. In one-shot MOO, the decision space typically consists of vectors of decision variables that describe, e.g., a candidate design. In contrast, a MORL solution is a function, that maps states to actions, often represented by a neural network \cite{gpipdalegre23}. The objective space in MORL can be constructed by calculating the expected returns of each policy. However, the decision space, comprising sequences of decisions predicted by e.g. a neural network for given states, is complex. Consequently, our initial step is to create the decision space, which we refer to as the Behavior Space as it reflects the  behavior of the policy. The goal is to summarize the behavior of each policy, enabling the calculation of similarity (or distance) between these summaries.

Techniques from Explainable RL can be particularly valuable in this context. Specifically, we are using Highlights \cite{Amir2018HIGHLIGHTSSA}, a policy summarization method, which identifies the most important states for the particular policy and states in the close proximity of those states. These are combined into summary videos of agent behaviors; details can be found in Section~\ref{policy_highlights}.
Other methods developed to summarize agent's behavior use computational models (\cite{huang2018establishing}, \cite{lage2019exploring}) or state abstraction (\cite{sreedharan2020tldr}, \cite{topin2019generation}). We decided not to use them as they produce summaries, which cannot be easily interpreted and analyzed for further computation of clusters.

\section{Approach}

\begin{figure}[!htb]
    \centering    
    \includegraphics[width=\columnwidth]{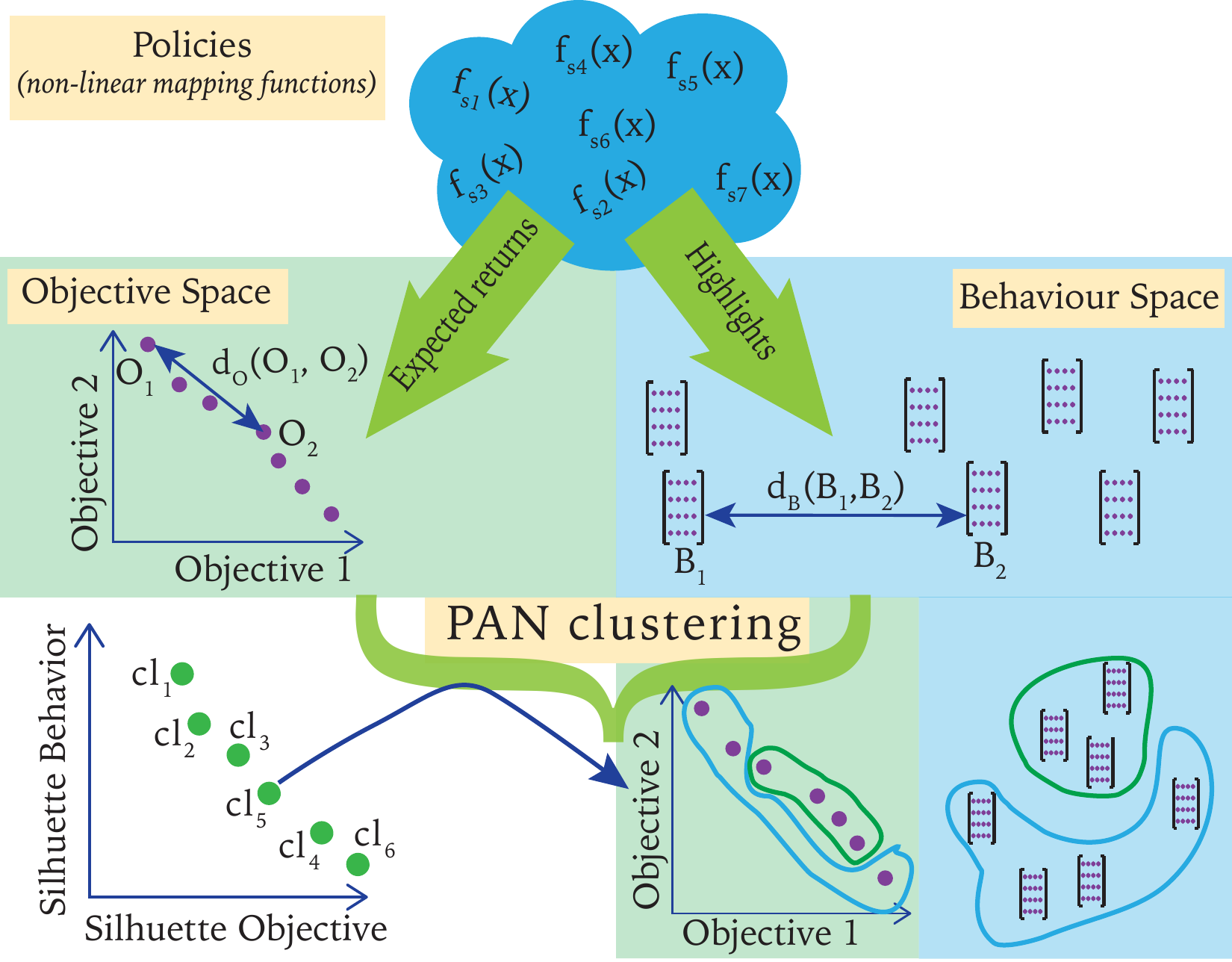} % first figure
    \caption{An outline for our approach which clusters a set of policies, considering clustering quality in both objective and behavior spaces.}
    %\caption{Outline of our approach to the problem of clustering of the solution set as a bi-objective problem. Given the solution set (non-linear mapping functions), the functions (policies or solutions) can be represented as vectors of objective values or highlights of their behavior. Given these two representations, PAN clustering can be applied to detect clusters in the Solution Set, which include information about the trade-offs in the objective space, as well as the behavior of the policies}
    \label{fig:process}
\end{figure}

Figure~\ref{fig:process}shows an overview of our approach, whose input is the set of policies generated by an MORL agent. Each policy, \(f_{i}(x)\), is a mapping function (states to actions), and is represented in two spaces. 

\begin{description}
    \item [Objective space:] An objective vector (vector of objective values), \(O_i\), computed as the expected returns over a number of episodes.
    \item [Behavior space:] A behavior matrix, \(B_i\), of five Highlights (Section~\ref{policy_highlights}) states.
\end{description}

We measure the distances between two policies in the objective and behavior spaces, \(d_O\) and \(d_B\), via Euclidean and Frobenius distances, respectively (Section~\ref{distances}).
These distances are used by PAN clustering (Section~\ref{pan}), a bi-objective clustering algorithm. PAN outputs a set of partitions (clustering), offering different trade-offs in terms of clustering quality in the two spaces. Each partition, \(C_i\), represents a clustering (set of clusters) of the policies.

The key advantage of this method is its flexibility---it allows for the customization of partitioning based on different preferences or requirements. For instance, a decision maker may prioritize the analysis of policy behavior over the optimization of objectives, or vice versa. This adaptability ensures that the clustering process aligns with the decision maker's goals, facilitating a more targeted and insightful exploration of the solution space.

\subsection{Policy Representation using Highlights}
\label{policy_highlights}
The Highlights algorithm \cite{Amir2018HIGHLIGHTSSA} is designed to create a video summary of an agent's behavior through online simulations. It calculates the most significant states encountered by the policy, including those nearby. The algorithm uses a concept of state importance, which is judged based on how critical the choice of action in that state is to future rewards, as indicated by the agent's Q-values. This method of determining importance has been effective in identifying teaching moments in student-teacher RL, although it does have limitations, such as sensitivity to the number of possible actions.

In the Highlights approach, the state importance was as:
\begin{equation}
I(s) = \max_{a}Q^\pi(s, a) \ - \min_{a}Q^\pi(s, a) \ ,
\label{importance}
\end{equation}
\noindent where \(Q_{(s, a)}^\pi\) quantifies the expected return of taking action \(a\) in state \(s\), following the policy \(\pi \) thereafter  \cite{10.5555/2484920.2485086}.
The original Highlights algorithm outputs videos of the most important transition for the policy.
It also includes transitions proceeding and preceding the important states so that these snapshots can be used to show to the users.

%How highlights are used
Utilizing the Highlights approach, we identify five most significant states for each policy (as in the original paper). We treat these five states as representative of policy behavior. Subsequently, each policy is represented by five states, which can be represented as a \(n_\text{feat} \times 5\) matrix, where \(n_\text{feat}\) is the number of features the state is represented with (we assume that it is possible to represent the state numerically). We also create video snapshots for later analysis.

%How we created the highights
We execute each policy fifty times to determine the set of five most important states for our analysis. During these runs, we assess the importance of each visited state based on Equation~\ref{importance}. We gather states from 50 executions of the policy, ensuring that each state is only recorded once by discarding any duplicates. 

\subsection{Distances between the Policies} 
\label{distances}

Our goal is to cluster policies, considering both objective and behavior spaces. To do so, we define a distance measure for each space. 

\begin{itemize}
    \item For the Objective Space, where policies are represented as objective vectors, we employ the Euclidean distance:
    
    \begin{equation}
        d_O(O_i, O_j) = \sqrt{\sum_{x=1}^{n_{\text{obj}}} (O_{i}^{x} - O_{j}^{x})^2},
    \end{equation}
    \noindent where \(O_i, O_i\) are objective vectors of policies \(i\) and \(j\), respectively; and \(n_{\text{obj}}\) is the number of objectives in the MORL problem.

    \item In the behavior space, each policy is represented by a matrix with each column corresponding to a state. To calculate the distance between two matrices, we use the Frobenius distance:
    
    \begin{equation}
        d_B(B_i, B_j) = \sqrt{\sum_{x=1}^{n_{\text{feat}}} \sum_{y=1}^{n_{\text{st}} = 5} (B_{i}^{x,y} - B_{j}^{x,y})^2},
    \end{equation}
    \noindent where \(B_i, B_j\) are behavior matrices of policies \(i\) and \(j\), respectively; \(n_{\text{feat}}\) is the number of features a state is represented with; and \(n_{\text{st}}\) is the number of states the policy is represented with (in our case, \(n_{\text{st}} = 5\)).
\end{itemize}

\subsection{Bi-Objective Clustering using PAN}
\label{pan}
% Shape of the behavior representation
In our case PAN considers both the objective values of a policy and its behavior, represented by the five Highlights states.

%silhuette index
PAN clustering requires a cluster \emph{validity index}, which measures how good the clusters are in a space. We employ the well-known Silhouette index \cite{rousseeuw1987silhouettes}. Given a clustering $C$ of $N$ data points, the Silhouette index is:

\begin{equation}
S(C) = \frac{1}{N} \sum_{i=1}^{N} s(i)
\end{equation}
\noindent where $s(i)$ is the Silhouette index for a single data point $i$ and is calculated as:
\begin{equation}
s(i) = \frac{b(i) - a(i)}{\max\{a(i), b(i)\}}
\end{equation}

\noindent where \(a(i)\) is the average distance from the \(i\) to the other data points in the same cluster, and \(b(i)\) is the minimum average distance from the \(i\) to data points in a different cluster, minimized over all clusters. \(S(C)\) ranges from -1 to 1, where a high value indicates that the clusters are cohesive and well-separated.

%the object is well matched to its own cluster and poorly matched to neighboring clusters.

The PAN algorithm seeks to minimize two objectives:

\begin{align}
\text{minimize} \quad & -(-S(C_{\text{O}}) - 1) \\
\text{minimize} \quad & -(-S(C_{\text{B}}) - 1),
\end{align}

\noindent where \(S(C_{\text{O}})\) and \(S(C_{\text{B}})\) are Silhouette indices, measuring the clustering performance in objective and behavior spaces, respectively.

%the representation

We represent a partitioning of the solution set in the PAN algorithm by a (variable length) list of clusters, where each cluster, in turn, is a list of indices of policies. Similar to the original paper, we also restrict that each partition contains at least two clusters and each cluster contains at least two policies. 

%How it works
The evolutionary algorithm starts with a random population of partitionings. Then, variation and selection operations are applied to the generated population as well as local optimization of the population in each generation (see \cite{ulrich-biobjective-2013}). The selection is conducted based on the hypervolume values, with reference point set to [2, 2] \footnote{The objectives, derived from the silhouette index, are transformed from a [-1, 1] range to [0, 2] for minimization.}

% parameters
As PAN operates as a standard evolutionary algorithm, it requires the configuration of parameters such as the number of generations, population size, and probabilities for various mutation operations. For every environment we tested the algorithm on, we did a parameter search over different combinations to achieve satisfactory performance. The parameter configuration as well as the convergence plots for each environment can be found in the appendix C \cite{appendix-our-paper}.

\subsection{Algorithm Overview}
Algorithm~\ref{algorithm} shows the psuedo-code of our approach. Its input is a set of policies \(F = \{f_{1}, \ldots, f_{i}\}\) and PAN parameters---the population size of the partitionings ($n$), number of generations to run the evolution for ($g$), and probabilities of mutation (\(p_m\)), the union of two randomly chosen clusters (\(p_u\)), recombination of parent clusters (\(p_r\)) and split of two randomly chosen clusters (\(p_m\)). 

First, the algorithm computes an objective vector for each policy as its expected returns and a behavior matrix for each policy using Highlights. Given these representations, the algorithm computes distances in objective and behavior spaces. Then, the algorithm performs PAN clustering. The final outputs a set of clusterings. Each clustering represents a set of clusters and the qualities of the clustering in objective and behavior spaces.

\begin{algorithm}[!htb]
\caption{Bi-objective clustering of a MORL solution set}
\begin{algorithmic}[1]
\State \textbf{Input:} \(F = \{f_{1,} \ldots, f_m\}\) \Comment{Solution set}
\State \textbf{Input:} $n$, $g$, \(p_m\), \(p_u\), \(p_r\), \(p_s\) \Comment{PAN parameters}

\State $S_O, S_B \gets [~]$ \Comment{Lists of objective vectors ($S_O$) and behavior matrices ($S_B$)}
%\State $S_B \gets [~]$ \Comment{List of behavior matrices}
\For{\(f_{i}\) in \(F\)}
    \State $S_O.\text{append}(\text{ExpectedReturns}(f_i))$
    \State $S_B.\text{append}(\text{Highlights}(f_i, 5))$
\EndFor
\State $D_O \gets \text{EuclideanDistances}(S_O)$ \Comment{Matrix of objective space distances}
\State $D_B \gets \text{FrobeniusDistances}(S_B)$ \Comment{Matrix of behavior space distances}
%\State $normBehDistances \gets \text{Normalize}(BehDistances)$
%\State $normObjDistances \gets \text{Normalize}(ObjDistances)$
\State $C \gets \text{PAN}(D_O, D_B, n, g, \text{\(p_m\), \(p_u\), \(p_r\), \(p_s\)})$
%\State \Return $setOfClusters$
\State \textbf{Output:} \(C = \{C_{1,} \ldots, C_n\}\) \Comment{Pareto set of clusterings; each $C_i$ is a set of clusters}
\end{algorithmic}
\label{algorithm}
\end{algorithm}

\section{Experimental Setting}

%\subsection{Multi-Objective Highway Environment}
We conduct experiments in four multi-objective environments from MO-Gymnasium  \cite{Alegre+2022bnaic}: MO-Mountaincar (3 objectives), MO-Minecart (3 objectives), MO-Highway (3 objectives), MO-Reacher (4 objectives). Appendix C \cite{appendix-our-paper} provides a detailed description of the environments, with convergence plots. 
For brevity, we deep dive into MO-Highway as a case study to demonstrate the applicability and insights offered by our algorithm. We utilize the MO-Highway environment because it closely mirrors real-world scenarios, making it comprehensible for end-users. Additionally, this environment contains only 12 policies within the solution set, enabling a thorough analysis and understanding of the behavior exhibited.

\begin{figure}[!htb]
    \centering
    \includegraphics[width=\columnwidth]{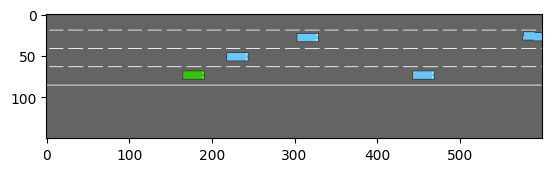} % first figure
    \caption{Screenshot from the highway environment\vspace{10pt}}%, where the ego-vehicle is green.} %\vspace{10pt} }
    \label{fig:highway}
\end{figure}

In the MO-Highway environment, an ego-vehicle (agent) is driving on a multi-lane highway populated with other vehicles (Figure~\ref{fig:highway}). The agent has three objectives---to reach a high speed, avoiding collisions with other vehicles, and driving on the right side of the road. The agent can move right or left (changing lanes), increase or decrease velocity, or stay idle, i.e., make no change. There is no defined target the agent is required to reach. Instead, the road goes on continuously. This property allows us to observe the agent’s general behavior and preferences instead of focusing on its progression towards reaching the goal. To represent the states, we use KinematicObservation (see \cite{highway-env} for details). It is a \(5 \times 5\) array that describes a list of 
5 nearby vehicles by a set of features of size 5.

% \subsubsection{Training the MORL agent}
% \label{subsec:training}
%GPI-PD and how the solution set was produced
To produce the solution sets for each environment, we sampled 1000 weight combinations, equally spaced in the weight simplex, using the Riesz s-Energy method \cite{rieszcardona20}. Given the weights and expected returns obtained by executing the policy for five episodes, the policies were compared and filtered based on their Pareto-dominance.  

%Performance of out agent
The MORL agents was trained with GPI-PD, using the same parameters and assumptions as in the MORL baselines \cite{felten_toolkit_2023} (default hyper-parameters, trained for 200k steps).

\section{Results}

%first, performance of all environments
First, we show the performance, measured via hypervolume of the resulting set of clusterings (or partitionings) for all four environments, followed by a detailed analysis of the MO-Highway environment.

%iterative k-medoid to 
To benchmark the performance of PAN, we compare the clusterings  identified by PAN with the clusterings identified via (iterative) $k$-medoids clustering \cite{schubert2019faster} (similar to the original paper \cite{ulrich-biobjective-2013}). The $k$-medoids algorithm is similar to the $k$-means algorithm. Whereas $k$-means requires data points to be represented in an $n$-dimensional (typically, Euclidean) space, $k$-medoids only requires a matrix of dissimilarities between data points. This is useful for our setting since it allows any type of policy representation to be used.

We apply $k$-medoids several times for all possible cluster numbers.
Each time, we cluster the solutions twice, once in
behavior space and once in objective space, and
check whether the optimized partitionings satisfy the
constraints (i.e. containing at least two clusters,
where each cluster contains at least two solutions).
For all partitionings that satisfy the constraints,
we calculate the silhouette index in decision
space and in objective space. Finally, to compare the
resulting population with PAN, we reduce the number
of achieved solutions to the population size used with
PAN, using PAN’s selection procedure. 
We do not consider applying k-medoids to the combined behavior-objective space as the meaning of such a distance would be unclear.

Table~\ref{table:perf_comparison} shows that PAN performs comparably to k-medoids in the MO-Reacher and MO-Minecart environments. This similarity in performance can be attributed to k-medoids'  focus on either behavior or objective space, potentially leading to better clustering outcomes that enhance the hypervolume by increasing the dispersion of the set of clusterings. In contrast, PAN tends to prioritize a balanced behavior-objectives performance, resulting in more moderate clusterings that do not significantly spread the set of clusters, thus achieving a lower hypervolume. However, in the other two environments evaluated, our method surpasses k-medoids, demonstrating its capability to achieve comparable or superior performance in a broader range of scenarios.

\begin{table}[!htb]
\centering
\begin{tabular}{lcc}
\toprule
\textbf{Environment} & \textbf{PAN} & \textbf{$k$-medoids} \\
\midrule
MO-Highway & 1.73 & 1.66 \\
MO-Reacher & 2.46 & 2.42 \\
MO-Minecart & 2.61 & 2.62 \\
MO-Lunar-lander & 2.45 & 1.75 \\
\bottomrule
\end{tabular}
\caption{Hypervolume achieved by PAN vs. $k$-medoids clustering.}
\label{table:perf_comparison}
\end{table}

\subsection{Detailed Analysis of MO-Highway}

\paragraph{Solution Set.}
% 12 policies and Addressing the small solution set
The solution set obtained from the MO-Highway environment consists of 12 policies. Although this is a smaller solution set than what is usually seen in real-world problems \cite{quinn2019control}, it is sufficient to illustrate the challenges associated with exploring potential policies.  After analyzing the expected returns of these policies (the trade-off plot can be found in the Appendix A \cite{appendix-our-paper}), it was evident that all the policies effectively mitigate collisions (one of the objectives). Thus, in the following analysis, we only focus on the trade-offs between two objectives---speed vs. staying on the right lane.

% First, k-medoids clustering
\paragraph{Objective and Behavior Spaces}

\begin{figure*}[!htbp]
    \centering
    \begin{subfigure}{0.19\textwidth}
        \centering
        \includegraphics[width=\linewidth]{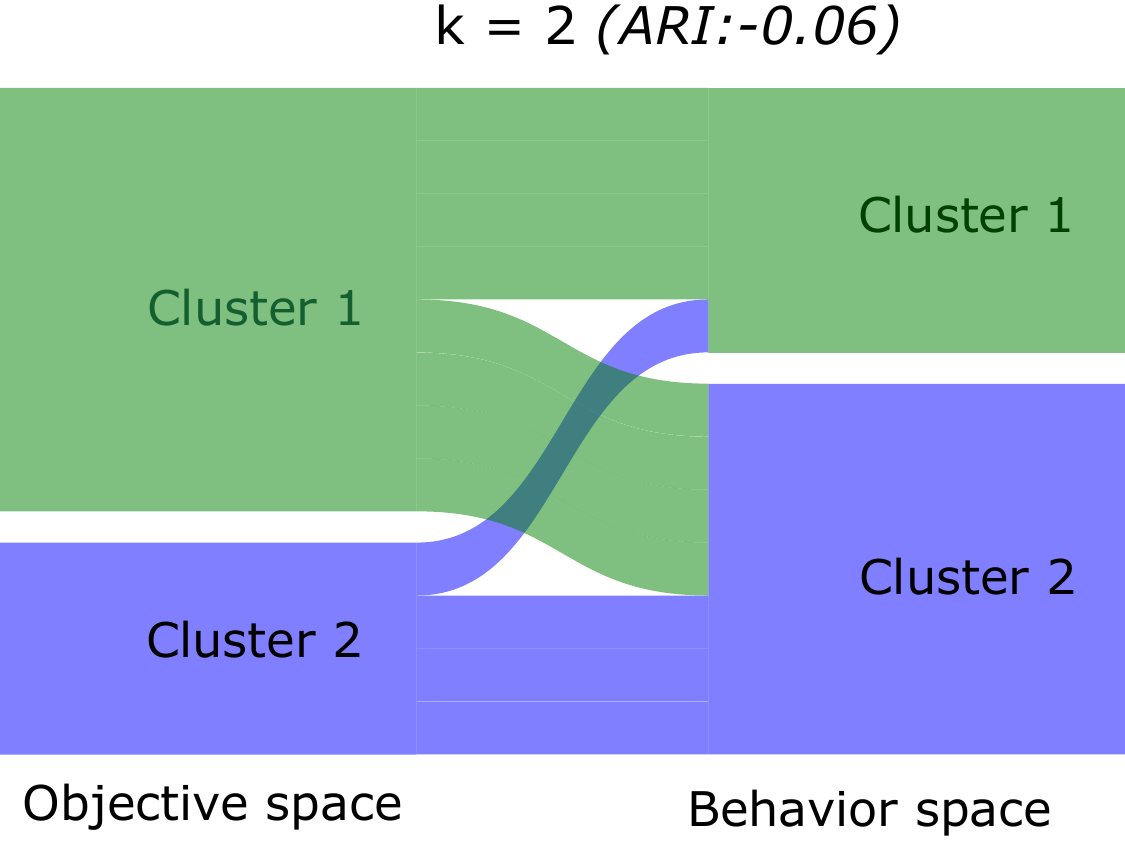}
        \label{fig:sankey_k2}
        \hfill

    \end{subfigure}
    \begin{subfigure}{0.19\textwidth}
        \centering
        \includegraphics[width=\linewidth]{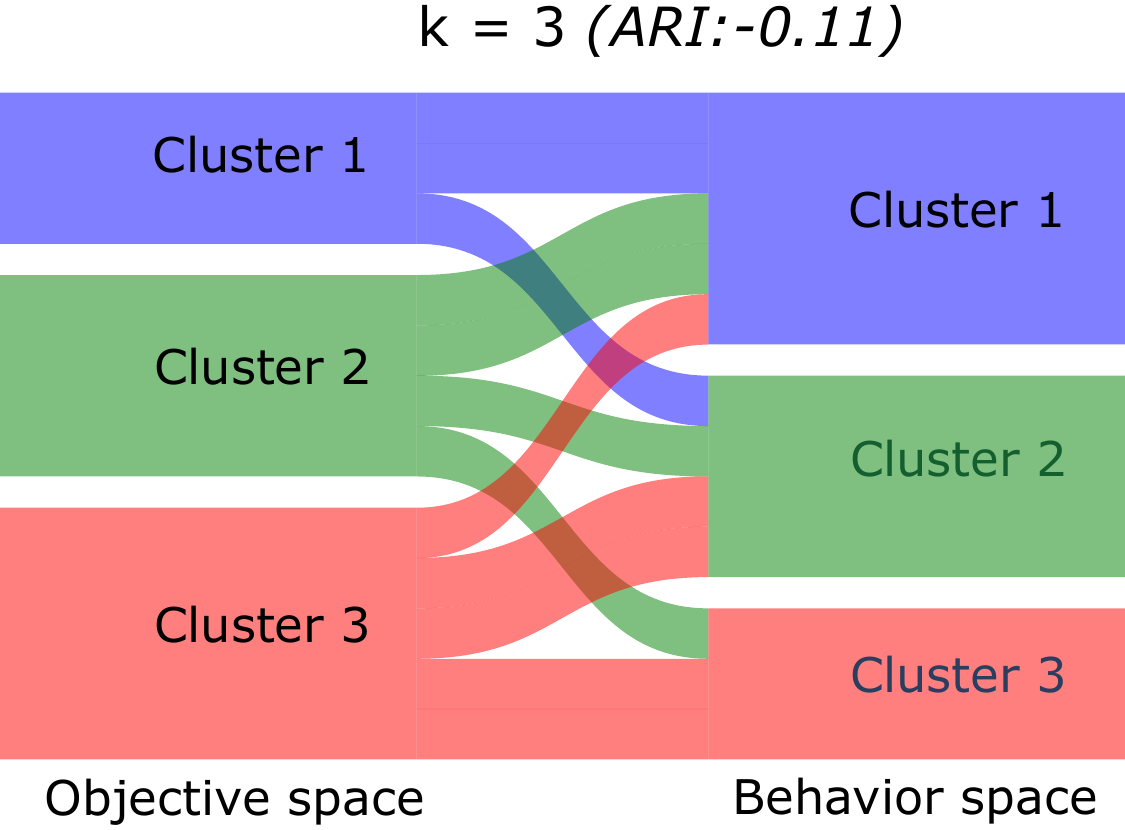}
        \label{fig:sankey_k3}
            \hfill

    \end{subfigure}
    \begin{subfigure}{0.19\textwidth}
        \centering
        \includegraphics[width=\linewidth]{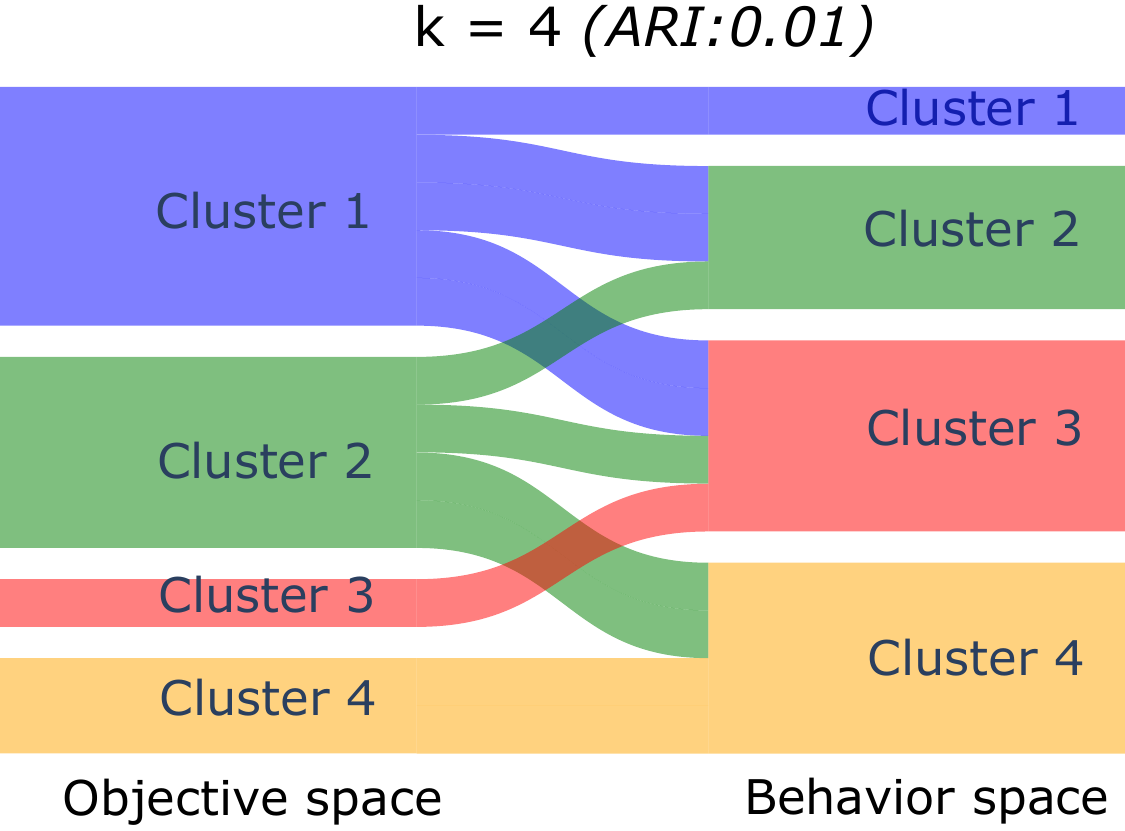}
        \label{fig:sankey_k4}
            \hfill

    \end{subfigure}
        \begin{subfigure}{0.19\textwidth}
        \centering
        \includegraphics[width=\linewidth]{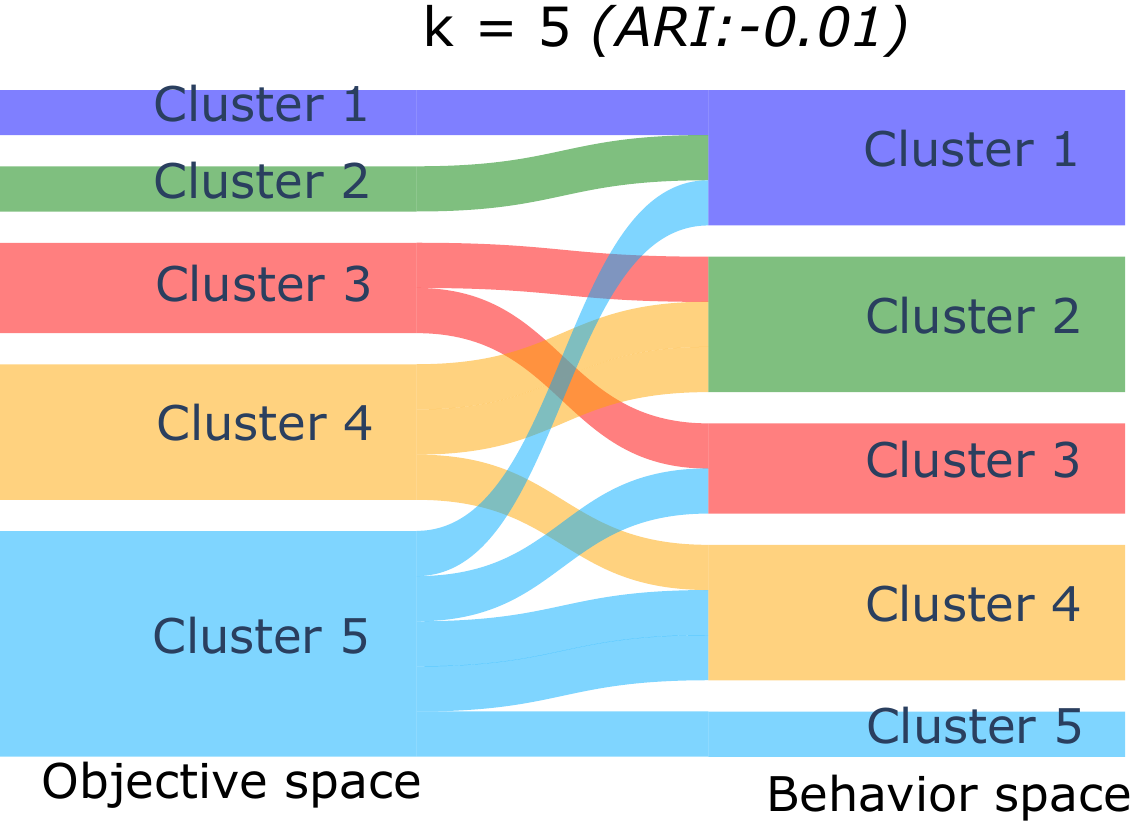}
        \label{fig:sankey_k5}
            \hfill

    \end{subfigure}
        \begin{subfigure}{0.19\textwidth}
        \centering
        \includegraphics[width=\linewidth]{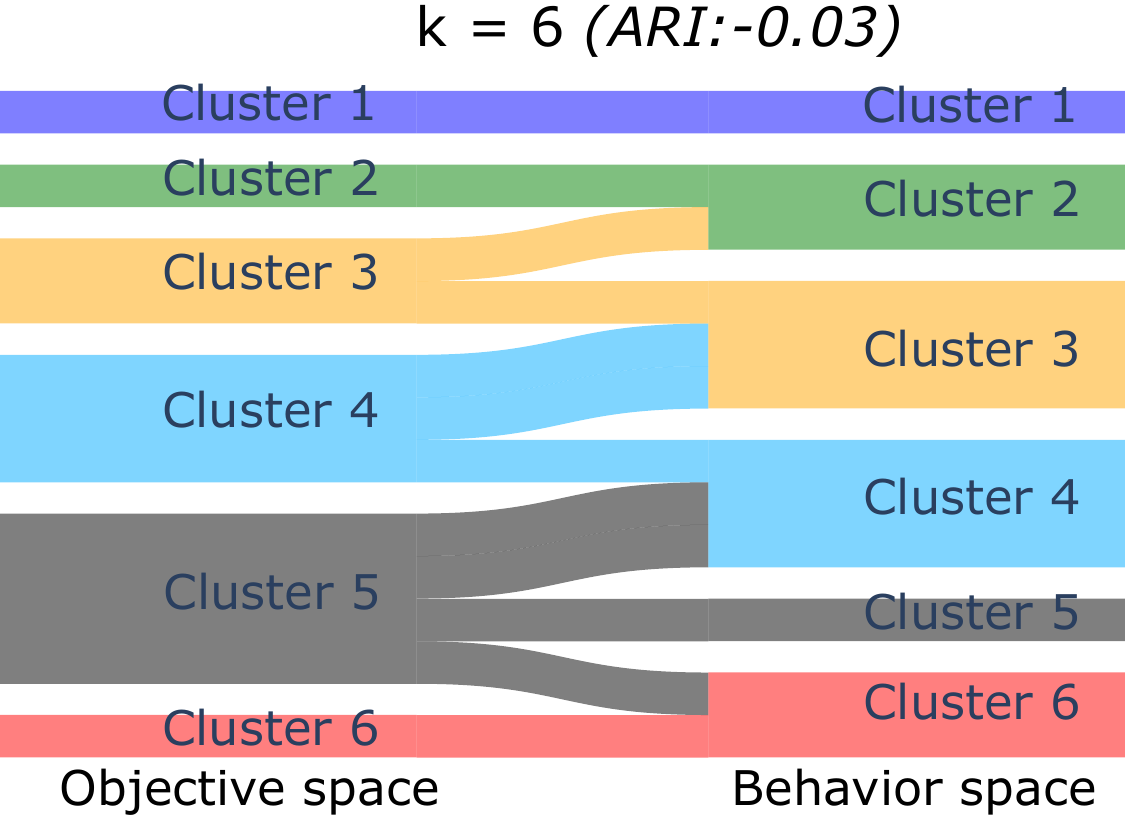}
        \label{fig:sankey_k6}
            \hfill

    \end{subfigure}

    \caption{Sankey diagram showing similarity between clusters in objective and behavior space. Each clusters is a node and the links are policies. \vspace{10pt} }
    \label{fig:sankey}
\end{figure*}

We cluster the solution set separately in the objective and behavior space to evaluate whether and to what extent the two sets of clusters differ using $k$-medoids clustering. Figure~\ref{fig:sankey} shows Sankey diagrams depicting the overlap between clusters in the two spaces. It is evident from these diagrams that the set of cluster differ substantially between the two spaces.

To quantify the difference between the two sets of clusters, we employ the Adjusted Rand Index (ARI). ARI adjusts the Rand Index, which measures similarity between two data clusterings by considering all pairs of samples and counting pairs that are assigned in the same or different clusters in the two clusterings. The values of the Adjusted Rand Index can range from -1 to 1, where 1 indicates perfect agreement between the two clusterings, 0 suggests random or chance agreement between the two clusterings, and a value less than 0 indicates that the observed clustering is less accurate than would be expected by random chance \cite{huber-comparing}. As Figure~\ref{fig:sankey} shows, most ARI values are close to zero. This demonstrates that the clustering in objective and behavior spaces can lead to very different sets of clusters.

\paragraph{PAN Clustering}

\begin{figure}[!htb]
    \centering
        \includegraphics[width=0.7\linewidth]{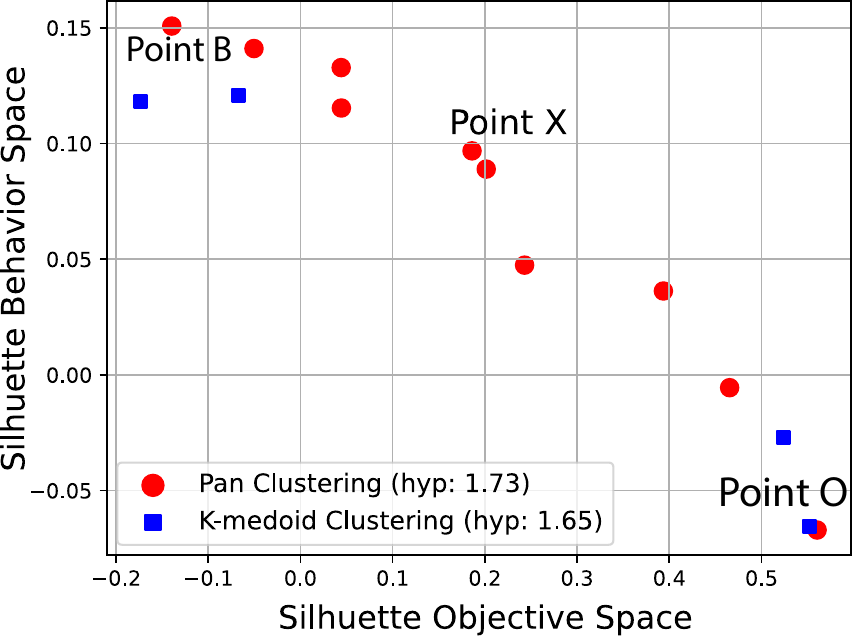} % first figure
        \caption{Clusterings obtained by the PAN clustering (red dots) and iterative k-medoids clustering (blue rectangles).\vspace{10pt} \vspace{10pt}
        } 
        \label{fig:clustering}
\end{figure}

Having shown that the clusters in objective and behavior spaces differ substantially, we report the results of PAN clustering, which considers both spaces. The ten red dots in Figure~\ref{fig:clustering} are the ten clusterings PAN found. Each of these clusterings considers a different trade-off between the clustering qualities (measured via Silhouette index) in the objective and behavior spaces.

As mentioned in the previous section, to benchmark the performance of PAN, we compare the clusterings identified by PAN with the clusterings identified via (iterative) $k$-medoids clusterings. Of the ten $k$-medoids clusterings we performed (Figure~\ref{fig:sankey}), we only consider four clusterings that adhere to the criteria of valid partitioning. The four blue rectangles in Figure~\ref{fig:clustering} represent the four $k$-medoids clusterings. First, from the figure, it is evident that the clusterings represented by blue points is either good in the objective space or the behavior space, but there are no compromise clusterings. In contrast, PAN offers a wider range of clusterings covering the full spectrum.

\begin{figure*}[!htbp]
    \centering
    \begin{subfigure}{0.32\textwidth}
        \centering
        \includegraphics[width=\linewidth]{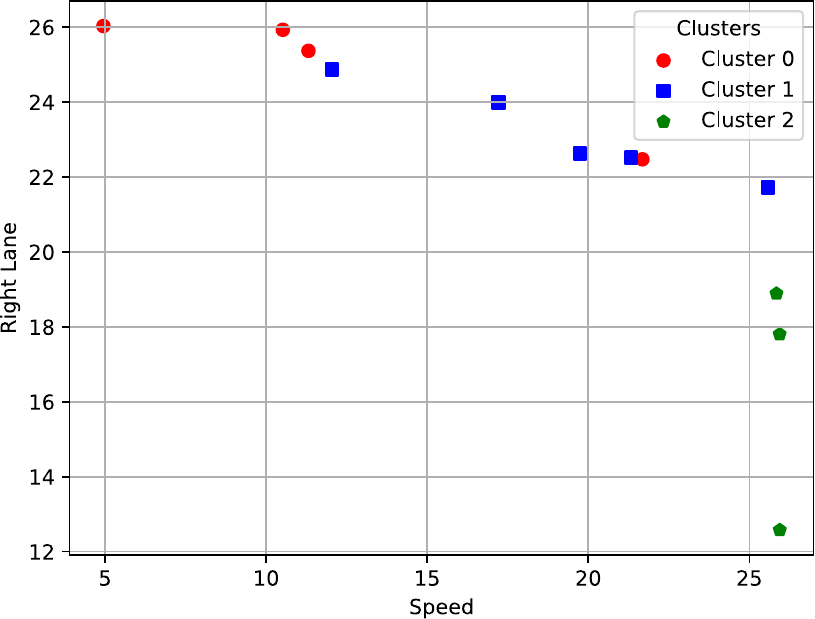}
        \caption{Clusters of the policies in the objective space based on the middle clustering (Point X).
        \vspace{10pt}}
        \label{fig:clustered}
    \end{subfigure}
    \hfill
    \begin{subfigure}{0.32\textwidth}
        \centering
        \includegraphics[width=\linewidth]{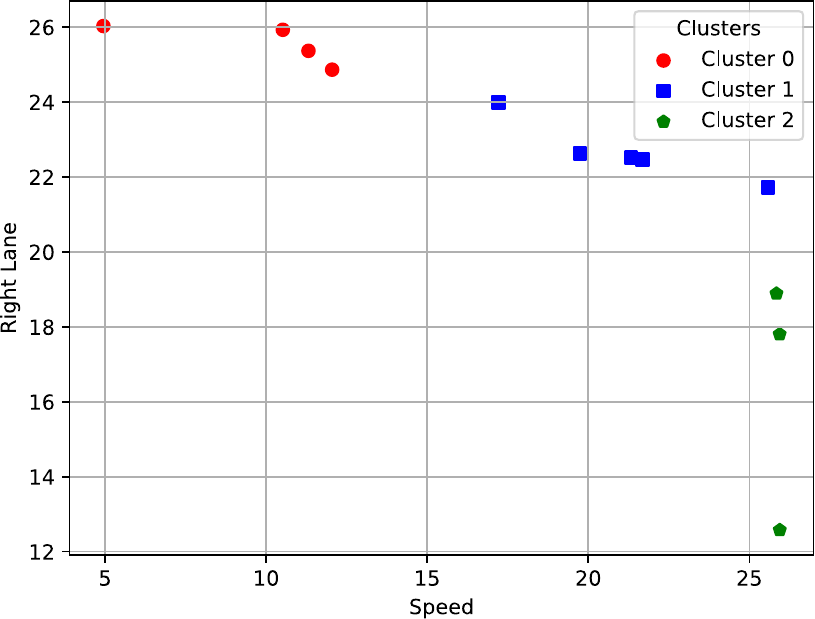}
        \caption{Clusters of the policies in the objective space based on the clustering with preference for objective values (Point O). \vspace{10pt}}
        \label{fig:clustered_obj}
    \end{subfigure}
    \hfill
    \begin{subfigure}{0.32\textwidth}
        \centering
        \includegraphics[width=\linewidth]{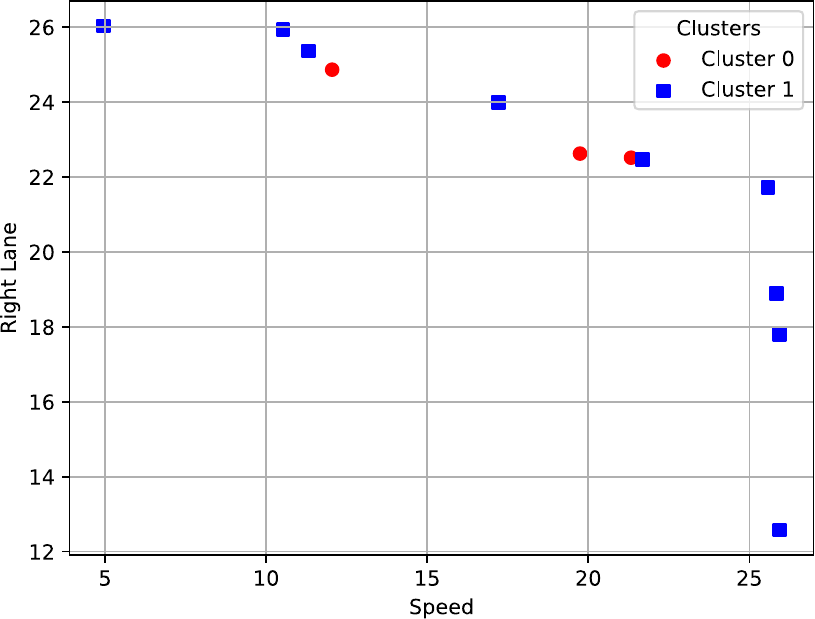}
        \caption{Clusters of the policies in the objective space based on the clustering with preference for behavior (Point B). \vspace{10pt}}
        \label{fig:clustered_beh}
    \end{subfigure}
    \caption{Clusters visualised in the objective space for chosen point in the Figure~\ref{fig:clustering} \vspace{10pt}}
    \label{fig:different_clust}
\end{figure*}

Each point in Figure~\ref{fig:clustering} represents a partitioning of the solution set. Figure~\ref{fig:different_clust} shows how three different partitionings look like in the Objective Space. These partitionings represent 1) compromise clustering, 2) clustering with preference for behavior, and 3) clustering with preference for objective values, represented by Points X, B and O, respectively, in Figure~\ref{fig:clustering}.
We show only two objectives (even though MO-Highway is a three-objective problem) as one of the objectives (Collision) does not offer any trade-offs. 

\begin{figure*}[!htbp]
    \centering
    \begin{subfigure}{0.32\textwidth}
        \centering
        \includegraphics[width=\linewidth]{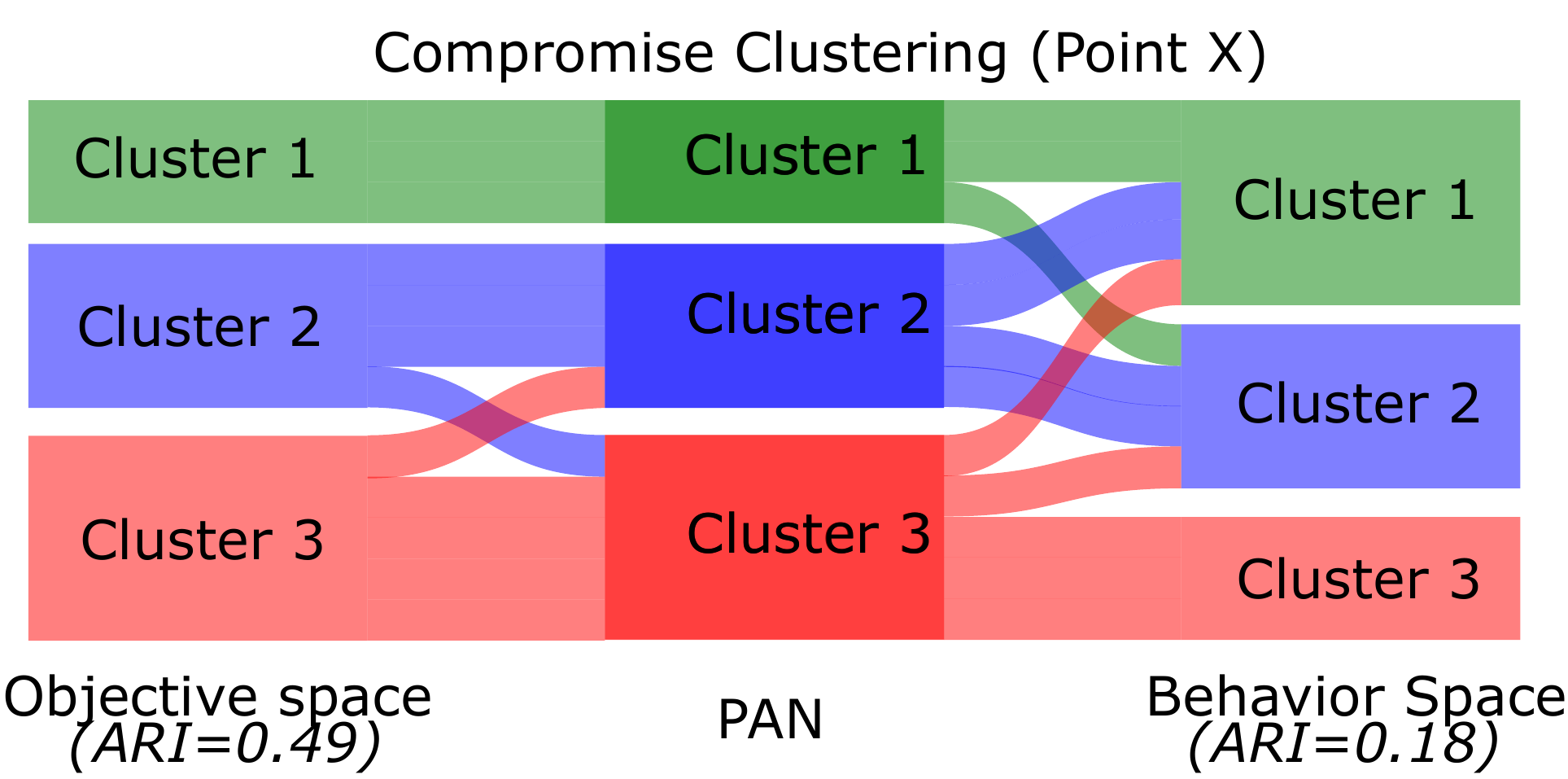}
        \label{fig:sankey_x}
    \end{subfigure}
    \hfill
    \begin{subfigure}{0.32\textwidth}
        \centering
        \includegraphics[width=\linewidth]{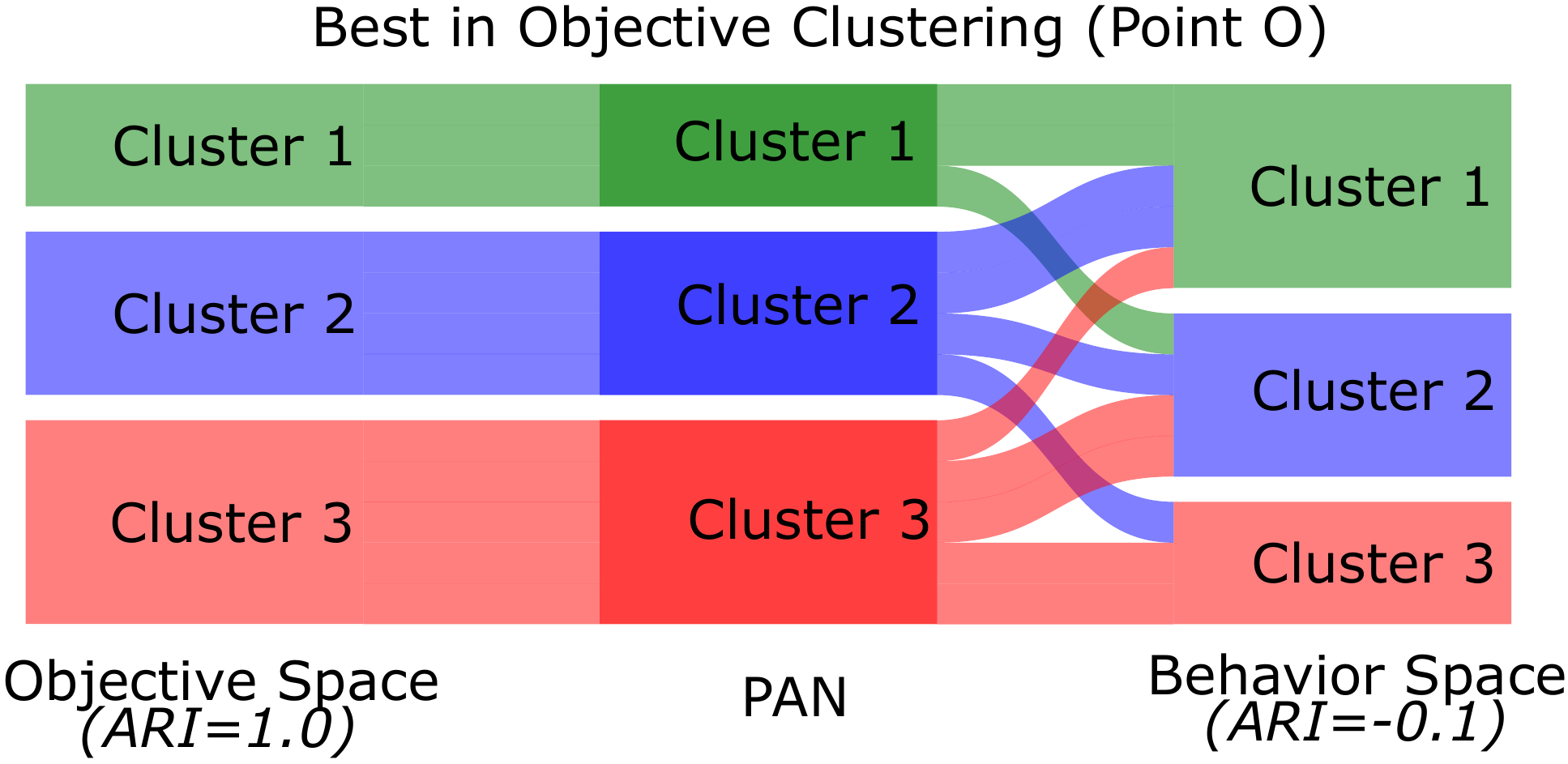}
        \label{fig:sankey_o}
    \end{subfigure}
    \hfill
    \begin{subfigure}{0.32\textwidth}
        \centering
        \includegraphics[width=\linewidth]{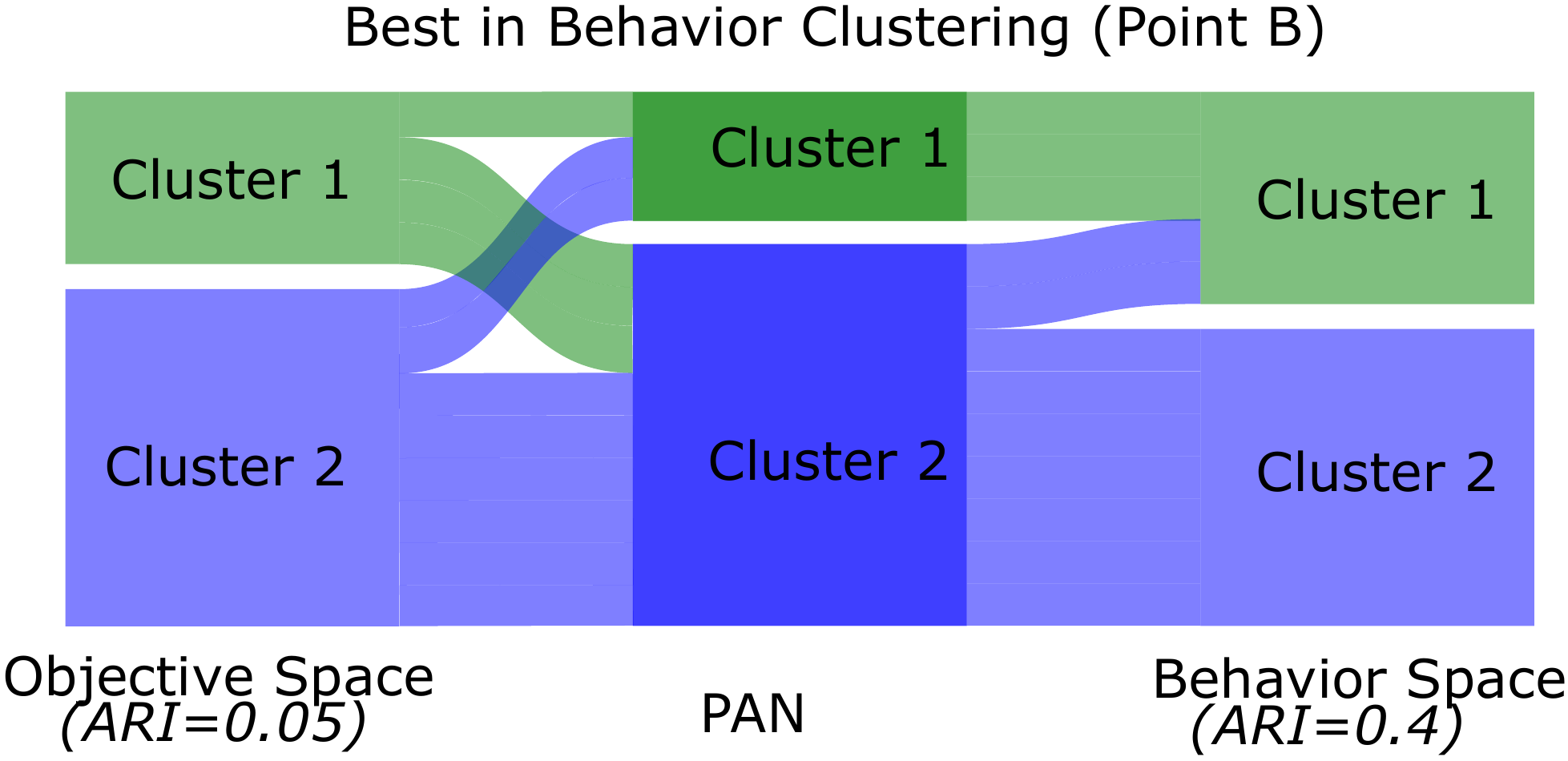}
        \label{fig:sankey_b}
    \end{subfigure}
    \caption{Sankey diagrams illustrating the correspondence between clusters generated by PAN and those identified through $k$-medoids clustering for three chosen PAN clusterings. Three points chosen are: a compromise clustering (point X from Figure~\ref{fig:clustering}) and two clusterings representing the extreme values---best clustering in the objective space and the best clustering in the behavior space (Point O and Point B). \vspace{10pt} }
    \label{fig:sankey_cl}
\end{figure*}

Finally, Figure~\ref{fig:sankey_cl} compares PAN clusterings with behavior-only and objective-only clusterings in three cases:%\vspace{-\baselineskip}

\begin{description}[nosep]
    \item [Compromise clustering] The top Sankey diagram shows a compromise clustering from PAN (Point X in Figure~\ref{fig:clustering}) in the middle, consisting of two clusters. The left and right columns show the clusters resulting from $k$-medoids clustering ($k = 3$) in objective and behavior spaces, respectively.
    \item [Clustering with preference for behavior] The bottom-left Sankey diagram shows a clustering from PAN which gives behavior the highest preference (Point B in Figure~\ref{fig:clustering}), consisting of two clusters, in the middle. The left and right columns are chosen from $k$-medoids clustering ($k = 2$) similar to the above case.
    \item [Clustering with preference for objective values] The bottom-right Sankey diagram shows a clustering from PAN which gives objectives the highest preference (Point O in Figure~\ref{fig:clustering}), consisting of three clusters, in the middle. The left and right columns are chosen from $k$-medoids clustering ($k = 3$) similar to the above cases.
\end{description}

Figure~\ref{fig:sankey_cl} also shows ARI values comparing PAN clustering with behavior space clustering, and PAN clustering with objective space clustering. From the diagrams and ARI values it is evident that choosing an extreme clustering from PAN aligns better with the corresponding space (objective/behavior space) than the other space (behavior/objective space). In contrast, the compromise clustering balances alignment with both spaces. Eventually, which of these clusterings to choose depends on the preference of the decision maker.

\section{Discussion}

We analyzed the snapshot videos produced by the highlights algorithm \cite{Amir2018HIGHLIGHTSSA}, and observed distinct policy behaviors---e.g., a tendency to decelerate during lane changes, maintaining a short distance between vehicles, and overtaking with marked caution (keeping too much distance between cars in front)---that are not apparent solely from the trade-offs in the objective space. For comparative analysis, we selected two clusterings to observe variations in cluster configurations, as illustrated in Figures~\ref{fig:clustered} and~\ref{fig:clustered_obj} visualized in the objective space.

Unsurprisingly, Figure~\ref{fig:clustered_obj} shows that the clusters are well-defined in the objective space for the given clustering. When comparing two clusterings (Point X and Point O), we observe differences in solution distribution solely between Cluster 0 and Cluster 1, while the policies in Cluster 2 remain consistent across both clustering options.

All three policies in Cluster 2 are notably fast and exhibit similar objective values. Upon reviewing the highlight videos, it is evident that two of these policies display similar behaviors, characterized by speed and flexibility during overtaking maneuvers. However, the third policy consistently maintains a significant distance from the preceding vehicle, indicating a more cautious approach to overtaking, which in the real-world can be dangerous. As Point X maintains the homogeneity of clusters in both spaces, this is acceptable. 

In Clusters 0 and 1, two policies (2 and 0) switch their cluster affiliations, reducing the homogeneity of the clusters in the objective space for Clustering X. 
%Policies 2 and 0 switch the clusters they belong to. 
Analysis of the highlight videos reveals that, for Clustering X, Cluster 1 comprises policies more inclined to overtake compared to Cluster 0. This suggests that while homogeneity in the objective space for Point X decreases, the categorization of policies based on their behavior improves.

By offering multiple clustering options, bi-objective clustering enables the selection of the most appropriate one for the problem. For instance, selecting a knee point partitioning---a point on the Pareto front balancing competing objectives where gains in one would mean disproportionate losses in another---ensures consideration of both spaces in cluster formation. %These clusters can then be presented to the DM. 
Alternatively, if the DM has a preference for the number of clusters or one space over another, a suitable clustering can be chosen accordingly while also conveying information about the representation's efficacy in both spaces. This approach not only enhances the DM's awareness but also provides comprehensive information to support informed and explainable decision-making.

 In our proposed approach, DMs are presented with clusters derived from compromise clustering. Each cluster is characterized by the average trade-offs and behavior patterns of the solutions it contains. These patterns can be illustrated through highlight videos of each cluster's medoid or by identifying significant features within the highlights of the solutions in each cluster.

For the MO-Highway, we have analyzed these patterns through video examination; however, a more automated method is feasible through the assessment of feature importance derived from highlight's state representations. We elucidate the distinct behaviors and effectiveness of separation for each cluster, empowering DMs with a comprehensive understanding of each cluster's characteristics.

Depending on the role of the DM---non-technical DM, technical DM (analyst), or developer---this method can be used for different purposes. For non-technical decision makers, we suggest displaying only representative policies from each cluster, like the cluster's medoid, to allow straightforward selections based on their preferences and avoid overwhelming them. They can then implement these directly or as a stochastic mixture of the cluster’s policies \cite{hayes2022practical}. For the technical DM, who is able to analyse solution set and understands how it was produced, it can help to identify trends within the clusters by analysing both spaces and the clusters. Additionally, regarding developers, this method holds the potential for enhancing interactive MORL algorithms, particularly during the preference-learning phase. By frequently soliciting DMs' preferences in an interactive manner, the algorithm can more accurately tailor its focus to areas of interest, thereby optimizing the learning process and outcomes.

\section{Conclusions and Future Directions}

We address post-analysis of MORL-generated multi-dimensional solution sets to elucidate trade-offs and policy behaviors for decision makers tasked with policy analysis and implementation. We employ clustering as a means of reducing the complexity of a solution set by developing a partitioning strategy effective in both behavior and objective spaces, optimizing for two indices related to these spaces. 

We demonstrate that effective partitioning in one space does not always correspond to similar results in the other, especially in MORL contexts where policies with different behaviors produce similar outcomes. We address the bi-objective challenge of partitioning across behavior and objective spaces by using PAN clustering to optimize two validity indices without merging them into a single measure, allowing decision makers to visualize and prioritize trade-offs between the two spaces for informed strategy selection.

% Next steps
We identify multiple avenues for future work. A critical next step is conducting user studies to validate the efficacy of the proposed methods as well as the best approach to identify representative clustering. These studies can investigate the utility of different clustering techniques by comparing their performance and user perception. Specifically, it is important to examine how well users can make informed decisions with the aid of additional information like policy behavior and trade-off highlights. In practice, this means presenting participants with problem descriptions and policies characterized by trade-offs, enriched with clustering and behavioral information, to evaluate which format improves decision-making effectiveness. This user-centric approach is pivotal in ensuring that the theoretical advancements in MORL translate into practical tools that enhance decision-making processes in real-world scenarios. Additional to that, we acknowledge that parameter choices for Highlights can be tuned specifically to each environment, which in turn, can yield better PAN clusterings
Another important direction is automating the analysis of highlight videos by detecting key features in highlighted states to enhance cluster descriptions, particularly as the number of policies grows. Finally, new methods for capturing policy behavior should be studied further. In this paper we focused on the clustering as a mean to reduce the solution set. However, there exist other approaches for redusing the size of the solution set, such as pruning methods \cite{PETCHROMPO2022108022}, which should be explored further.

\end{document}